\documentclass{article}
\usepackage{spconf,amsmath,graphicx}
\usepackage{hyperref}
\usepackage[acronym,toc]{glossaries}
\loadglsentries[\acronymtype]
\makeglossaries
\usepackage{graphicx}
\usepackage{xspace}
\usepackage{amsmath,amssymb} %
\usepackage{float}
\usepackage{booktabs}
\usepackage{color}
\usepackage{balance}

\newcommand{\prob}{\mathbb{P}}

\newcommand{\dst}[1]{\mathcal{I}_{\mathrm{#1}}}
\newcommand{\suphash}{SSH\xspace}
\newcommand{\suphashfull}{SH\xspace}
\newcommand{\imagenet}{ImageNet\xspace}
\newcommand{\map}{mAP\xspace}

\newacronym{CNN}{CNN}{Convolutional Neural Network}
\newacronym{PQ}{PQ}{Product Quantizer}

\title{How should we evaluate supervised hashing?}
\name{Alexandre Sablayrolles \qquad Matthijs Douze \qquad Nicolas Usunier \qquad Herv\'{e} J\'{e}gou}
\address{\vspace{-10pt} Facebook AI Research}
\begin{document}
\maketitle
\begin{abstract}
{%
Hashing produces compact representations for documents, to perform tasks like classification or retrieval based on these short codes. 
When hashing is supervised, the codes are trained using labels on the training data.
This paper first shows that the evaluation protocols used in the literature for supervised hashing are not satisfactory: we show that a trivial solution that encodes the output of a classifier significantly outperforms existing supervised or semi-supervised methods, while using much shorter codes. 
We then propose two alternative protocols for supervised hashing: one based on retrieval on a disjoint set of classes, and another based on transfer learning to new classes. 
We provide two baseline methods for image-related tasks to assess the performance of (semi-)supervised hashing: without coding and with unsupervised codes. These baselines give a lower- and upper-bound on the performance of a supervised hashing scheme. }
\end{abstract}

\section{Introduction}

Traditional hashing aims at reproducing a target metric, such as the cosine similarity, based on compact codes. Large databases are then be stored memory and queried efficiently. Algorithms are tuned on a \textit{learn} set, and evaluation is done by looking for nearest neighbors of a \textit{query} set in a \textit{database}.

Recent works have proposed to use annotated data, in the form of labeled images, 
to improve the hashing quality. Indeed, even if hash codes yield noisy reconstructed vectors, being able to discriminate classes from these reconstructions is a desirable property. In the literature, the proposed evaluation protocol for this property involves two datasets: a \textit{train} set, with known or partially known labels and a \textit{query} set with unknown labels. The true positives are defined as images from the train set belonging to the same class as the query. This setting will be referred to as semi-supervised hashing (\textbf{\suphash}).

In the particular case of supervised hashing (\textbf{\suphashfull}),  the labels of the train set are known. We can train a classifier and use it to classify queries, returning for each query all elements of the assigned class. This trivial baseline, which is not considered in overlooked in most works published on the topic, outperforms state-of-the-art methods. This shows that the evaluation protocol is flawed: it only requires to discriminate between known classes and not to reconstruct vectors in a semantically meaningful way. 

This paper makes the following contributions:
\begin{list}{$\circ$}{}
\item We show that both \suphashfull and \suphash are well addressed by a trivial encoding on the output of a classifier, which outperforms the results reported in the literature. 
\item We propose two tasks and corresponding baselines to assess the performance of (semi-)supervised hashing: transfer to retrieval and transfer learning. They %
correspond to real use-cases.
\item We show that, in the case of transfer learning, it is better to insert the layer producing a compact code in the middle of the network. In contrast, existing methods routinely encode the last activation layer. 
\end{list}

\section{Related work}
\label{sec:related}

We distinguish three classes of methods for supervised hashing: triplet loss hashing~\cite{NFS12,WLSJ13,WWYL13,ZHWT15}, pairwise similarity-based and label-based. 
Often, pairwise similarity and label information are equivalent, because pairwise similarity is defined as sharing the same label, and reciprocally labels are equivalence classes of pairwise similarity relations. However they are treated differently because constructing pairwise similarity matrices scales quadratically with the number of labeled samples, limiting these algorithms to small labeled sets. 
\medskip
 
\noindent \textbf{Pairwise similarity based.} Binary Reconstructive Embeddings (BRE) \cite{KD09} minimize the distortion between the distance matrix in the original space and the Hamming distances of the codes. BRE is extended to the supervised case by replacing the distance matrix by a pairwise similarity matrix. Following this work, different techniques %
use both the similarity matrix and the feature space: semi-supervised hashing~\cite{WKC10}, supervised hashing with kernels (KSH)~\cite{LWJJC12}, Semantic Hashing~\cite{SH09}, Minimal Loss Hashing~\cite{NF11}, fast supervised hashing \cite{LSSHS14}. 
\medskip

\noindent \textbf{Label based.} Supervised discrete hashing (SDH) \cite{SSLT15} and supervised quantization (SQ)\cite{WZGTW16} integrate the labels in a classification loss, along with a hashing loss. Recent work also explore deep architectures \cite{DDC16, ZCS16} and augmented Lagrangian \cite{DDNC16} for supervised hashing.
\medskip

\noindent \textbf{Transfer learning.}
Indexing based on attributes or unrelated classes is standard~\cite{LNH09,TSF10}. Torresani et al.~\cite{TSF10}  remark that, ``Without the novel-category requirement, the problem is
trivial: the search results can be precomputed by running the known category
detector on each database image [...] %
and storing the results as
inverted files''. 
We apply this remark to hashing and extend it to the semi-supervised setting.

\section{Supervised hashing: a simple baseline} %

\newcommand{\cl}{\textrm{cl}}
\newcommand{\AP}{\textrm{AP}}

This section describes the protocols used in the literature for \suphash and \suphashfull, 
and discuss how a simple strategy efficiently solves the corresponding problems. 

\subsection{Evaluation protocols of \suphash and \suphashfull}

The task of \suphash consists in indexing a dataset of $N$ images $
\dst{train}$, of which a subset $ \dst{label} \subseteq \dst{train} $
is labeled. \suphashfull is the extreme case $\dst{label} =
\dst{train}$. Given an unlabeled query image $q$, the system must
return an ordered list of images from the $\dst{train}$. For
evaluation purposes, a dataset of queries is given; the labels of the
queries as well as all labels in $\dst{train}$ are known to the
evaluator, even in the \suphash setting, and an image is deemed correct if it has the same label as the query. The performance is measured in terms of precision or mean average
precision (\map), which we now describe.

Given a query $q$, we first define $\delta(q, i) = 1$ if the $i$\textsuperscript{th}
image is correct for $q$, and $0$ otherwise. The precision at (rank)
$k$ is given by $P(q,k) = \frac{1}{k}\sum_{i=1}^k\delta(q,
i)$. Denoting by $\cl(q) = \sum_{i=1}^N \delta(q,i)$ the total number
of correct images in $\dst{train}$, %
the average precision at $k$ is $ \AP(q, k) = \frac{1}{\cl(q)}
\sum_{i=1}^k \delta(q, i) P(q, i)$. The \map at $k$ (or simply \map when $k=N$) is the mean $\AP$ over all
test queries.

\subsection{Retrieval through class probability estimation}

It is well known in information retrieval~\cite{R77} and learning to rank that
the optimal prediction for precision at $k$ is given by ranking items $x \in \dst{train}$
according to their probability of being correct for the query. This result extends to the optimization of mAP. 
\medskip 

\noindent \textbf{Optimal ranking for \suphashfull.}
In the specific setup of \suphashfull where the system knows the
labels of the images in $\dst{train}$, the probability that an image
$x$ with label $y$ is correct is the probability $\prob(y|q)$ that the query image
has label $y$. The important point here is that the probability of $x$ being correct for $q$ only depends on
the label of $x$. Thus, ordering the $C$ labels so that
$\prob(c_1|q) \geq ... \geq \prob(c_C|q)$, %
the \emph{optimal} ranking is to return 
all images of $\dst{train}$ with label $c_1$ first, followed by
all images with label $c_2$, and so on.

In practice, $\prob(.|q)$ is unknown, but
we can train a classifier on $\dst{label}=\dst{train}$ which outputs probability
estimates $\hat{\prob}(c|q)$ for every label $c$, and compute the
optimal ranking according to $\hat{\prob}(.|q)$. Such probability
estimates are given by, \emph{e.g.}, multiclass logistic regression or a
\gls{CNN} with a softmax output layer.
Labels of $\dst{train}$ are stored on $\lceil\log_2(C) \rceil$ bits or in an inverted file~\cite{TSF10}.
\medskip 

\noindent \textbf{Relationship between classification accuracy and ranking performance.}
If the classifier has classification accuracy $p$, then the
resulting \map is \emph{at least} $p$: whenever the classifier
correctly predicts the class of $q$, all images of that class will be
ranked first and the resulting $\AP(q)$ is $1$; this happens on a
proportion $p$ of the queries. Thus the classification accuracy is a lower bound on the \map.
\medskip 

\noindent \textbf{Optimal ranking for \suphash. }
In the more general setup of \suphash, we do not know the label of
some images in $\dst{train}$. Yet, considering the (true) conditional
label probabilities $\prob(c|q)$ and $\prob(c|x)$, the probability
that $x$ is correct for $q$ is given by $\sum_{c=1}^C
\prob(c|q)\prob(c|x)$: it is the probability that both $q$ and $x$
have the same label, assuming conditional independence of the labels of the
query and the image. Notice that this is the dot product between the
conditional label probability vectors of $q$ and $x$. Then, given
probability estimates $\hat{\prob}$ for the labels of queries and
images, which are obtained on $\dst{label}$, we consider two
retrieval algorithms:
\begin{description}
\item[\emph{Classifier topline:}] For each image $x$ of $\dst{train}$, store
  a vector $u(x)$ equal to either (1) the one-hot encoding vector of
  the label of $x$ if $x\in \dst{label}$, or (2) the full conditional
  probability vector $\hat{\prob}(.|x)$. Rank images $x$ according to
  the dot product $\left<\hat{\prob}(.|q), u(x)\right>$. This strategy
  corresponds to the optimal strategy, but requires storing the 
  probability vectors for images in $\dst{train} \backslash
  \dst{label}$.
\item[\emph{Classifier hashed:}] Here we hash the conditional probability vector. The first hashing method that we evaluate, is the one-hot strategy, which stores the index of the maximal activation on $\lceil\log_2(C) \rceil$ bits. This approach, denoted \textbf{Classifier+one-hot} in what follows, returns all images of the strongest class first. The second encoding, referred to as \textbf{Classifier+LSH}, is locality-sensitive hashing (LSH) with tight frames~\cite{JFF12}, a simple non data-adaptive hashing scheme. This LSH method produces binary vectors that are compared with Hamming distances. Therefore it can be used as drop-in replacements for the competing binary encoding methods.
\end{description}

\section{Experiments: classifiers on \suphashfull and \suphash}

Here we experiment with the classifier based retrieval method on \suphash. We use  off-the-shelf classifiers, whose accuracies are not necessarily the current state of the art. However, we show that they perform better than \suphash methods of the literature. %
We consider two datasets: CIFAR10 and \imagenet. 
\medskip

\noindent \textbf{CIFAR10}~\cite{K09} is a dataset of 32x32 color natural images that contains $60,000$ images divided into 10 classes. 
For deep methods, we compare against DSH in the \suphashfull setting, and use the provided train-test split of CIFAR10 to train an AlexNet. For non-deep methods, we follow the GIST-based protocol of \cite{LWJJC12,SSLT15,WZGTW16}.  We hold out $1,000$ query images ($100$ per class) and index the remaining $59,000$ images, a variable number $n_\textrm{label}$ of them being labelled (following the experimental protocols of the papers we compare with). 

We start from the 512D GIST descriptors of the images; then we sample $h$ of the labeled images $ (a_i)_{i=1}^h $ as anchors. Images are represented by their Gaussian features $\big[ \exp(-\| x - a_i \|_2^2 / 2 \sigma^2) \big]_{i=1..h}\in \mathbb{R}^h$, with $ \sigma = \frac{1}{N} \sum_{i=1}^N \min_{j=1, \dots, h} || x_i - a_j ||_2 $. We fit a Logistic Regression classifier on these features. We cross-validate the regularization parameter on 10\% of the train set.

\begin{table}[t]
\caption{\label{tbl:classif_cifar}Retrieval (mAP): \textsc{Cifar10}, SH and SSH protocols.}
\centering
{\small \noindent\begin{tabular}{*{6}{l}}  
\toprule
Features  & \textbf{$n_\textrm{label}$} & {$h$} & Method  & bits  & \map \\ \midrule
GIST & 59,000 & 1,000 & SQ~\cite{WZGTW16}   & 64 & 70.4 \\
(SH) &        &       & VDSH~\cite{ZCS16}        & 16 & 65.0 \\ %
 &&& SQ~\cite{WZGTW16}   &128 & 71.2 \\ 
&&& Ours+one-hot & \textbf{4} & \textbf{73.0} $\pm$ 0.5 \\ \midrule %
GIST & 5,000 & 1,000 &SDH~\cite{SSLT15} & 64 & 40.2\\ %
(SSH) &&& Ours+one-hot & 4 & 36.7 $\pm$ 0.5 \\    %
&&& Ours+LSH & 64 & \textbf{41.8} $\pm$ 0.7 \\
&&& Ours topline & - & \textit{47.7} $\pm$ 0.7 \\ \midrule
GIST & 1,000 & 300 & KSH~\cite{LWJJC12} & 12 & 23.2  \\ %
(SSH) &&& KSH~\cite{LWJJC12} & 48 & 28.4 \\ 
&&& Ours+one-hot & 4 & 27.0 $\pm$ 0.7 \\ %
&&& Ours+LSH & 48 & \textbf{31.0} $\pm$ 0.8 \\
&&& Ours topline & - & \textit{35.2} $\pm$ 0.8  \\
\midrule
Deep & 50,000 &- & DSH~\cite{LWSC16} & 12 & 61.6  \\  	
(SH) & && DSH~\cite{LWSC16} & 48 & 62.1 \\ %
AlexNet & && Ours+one-hot & \textbf{4} & \textbf{87.0} \\ 
\bottomrule
\end{tabular}
\vspace{-15pt}}
\end{table}
Results are shown in Table \ref{tbl:classif_cifar}. We compare our approach to methods in the literature, using the numbers reported in the cited papers. With $\lceil \log_2 (C) \rceil $ bits, \map results are almost as good as the state of the art, while being 4-8 times more compact. With the same code size, simple LSH encodings outperform competing methods by a large margin.

Although the work of \cite{XPLLY14} was both deep and a SSH setup, their evaluation metric differs from our definition of \map and thus we have not included them in our comparison.
\medskip

\noindent \textbf{\imagenet} (ILSVRC 2012), %
contains over $1.2$ million natural images of $1,000$ categories~\cite{DSLLF09}. The training set is used for learning and indexing purposes. The $50,000$ validation images are used as queries. In \cite{WZGTW16}, the images are represented by activations of a VGG16 %
network~\cite{SZ14}.

The authors of~\cite{WZGTW16}  experiment on \imagenet, in the \suphashfull setting. They use a \gls{CNN} trained on \imagenet, and then use the train labels to train their quantization method. We use the same classifier from Caffe (VGG16) to classify query images, and store train labels as 1-hot vectors. The results in Table~\ref{tbl:classif_imgnet} show that our baseline method is more accurate and an order of magnitude more compact.

\begin{table}[h]
\caption{\label{tbl:classif_imgnet}Results on ImageNet with the SH protocol.}
\centering {\small
\noindent\begin{tabular}{*{6}{l}}  
\toprule
Descriptors & Method  & bits  & \map@1500 \\ \midrule
VGG & SQ~\cite{WZGTW16} & 128 & 62.0 \\
VGG & Classifier+one-hot & \textbf{10} & \textbf{66.4} \\
\bottomrule
\end{tabular}}
\end{table}

\section{Proposed evaluation protocols}

The previous section has shown that existing protocols fail to capture desirable properties of supervised hashing schemes. In this section, we propose two evaluation tasks, namely retrieval of unseen classes, and transfer learning to new classes. They correspond to application cases on large and growing user-generated datasets, where classifiers are trained on fluctuating training sets or for new labels. For computational reasons the features cannot be recomputed when the classes evolve, and mid-level features must be compressed in a way that preserves their semantic information. The two protocols we consider differ only in the evaluation metric: ranking versus class accuracy. 
\medskip

\noindent \textbf{Dataset definition.}
In both tasks, we start from a standard classification dataset but 
we use separate classes at test time, similar to a metric learning setup~\cite{MVPC12}. 
75\% of the classes are assumed to be known when learning the hashing function, and the 25\% remaining classes are used to evaluate the encoding / hashing scheme. We call train75/test75 the train/test images of the 75\% classes and train25/test25 the remaining ones.

In practice, we shuffle the classes randomly and use 4 folds to define 4 of those splits. Performance measures are averaged over the folds.
We split the classes of an existing dataset rather than combining different datasets, because the latter would introduce a lot of noise due to dataset bias~\cite{TE11}. In both protocols, test75 is not used at all. 

As feature representations, we use the activation maps at a given level of a CNN trained on train75\footnote{Other types of features, such as GIST are also possible.}. The top-line is when these activation maps are stored completely. To evaluate the hashing, they are encoded using hashing methods that can reconstruct an approximation of the original features.

\medskip

\noindent \textbf{Protocol 1: Retrieval of unseen classes.}
In this setup, we use the hashing scheme to index train25 and use test25 as queries. For each query from test25, we retrieve nearest neighbors among train25 and then compute the \map. The nearest neighbors are defined by the L2 distance between descriptors. This is a relevant distance measure for CNN activation maps~\cite{BSCL14}.
The labels of train25 are used for evaluation only. This setup is like an instance search approach except that the ground-truth is given by the class labels. The train75 - train25 - test25 split is the supervised equivalent of the \textit{learn} - \textit{database} - \textit{query} split in unsupervised hashing. 
\medskip

\noindent \textbf{Protocol 2: Transfer learning. }
A new classifier with the same structure as the top of the original CNN is trained from scratch using the stored train25 descriptors.
The classification accuracy is reported on test25. The goal is to maximize the transfer accuracy on test25.

Compared to recomputing the features from the images, this approach offers two advantages. First, the features are stored in a compact and semantic way. Secondly, it avoids the  computationally intensive computation of the low-level activations: in the case of AlexNet on CIFAR-10, $80\%$ of the computation time is spent in the lower convolutional layers.

\section{Baselines on the transfer-based evaluation protocol}

\begin{table}[t]
\centering
    \caption{\label{tab:class_retrieval}Retrieval performance (mAP, Protocol~1), when features are extracted at different layers in the network. %
} \smallskip

\centering 
CIFAR-100 \medskip

{\small
\begin{tabular}{*{6}{r}}
\toprule
Layer & conv3 & fc1 & fc2 & fc3 & softmax\\
\midrule
Full & 15.6\% & 16.3\% & 21.3\% & 22.2\% & 22.8\% \\
PQ, M=4 & 16.6\% & 16.8\% & 21.0\% & 21.2\% & {\bf 22.0\%}  \\
\bottomrule
\end{tabular}}

\medskip

ImageNet \smallskip

{\small
\begin{tabular}{*{6}{r}}
\toprule
Layer    & fc1 & fc2 & fc3 & softmax & +PQ, M=8\\
\midrule
mAP & 4.72\% & 10.89\% & 11.3\% & 13.52\% & 11.4\% \\
\bottomrule
\end{tabular}}

\end{table}

\begin{table}[t]
\centering
\caption{\label{tab:transfer_accuracy}Accuracy in transfer (protocol 2) for CIFAR-100 when features are extracted at different layers. %
}
{\small
\begin{tabular}{*{4}{r}}
\toprule
Layer & conv3 & fc1 & fc2 \\
\midrule
Full & 69.6\% & 61.8\% & 57.7\%  \\
PQ, M=4 & 43.4\% & 45.3\% & 47.4\% \\
\bottomrule
\end{tabular}}
\end{table}

\begin{figure}
      \caption{Accuracy in transfer (Protocol 2), for layer conv3 on CIFAR-100, as a function of the number of bytes per image.  %
      \label{fig:transfer_accuracy_conv3}
}
\begin{center}
    \includegraphics[width=\linewidth]{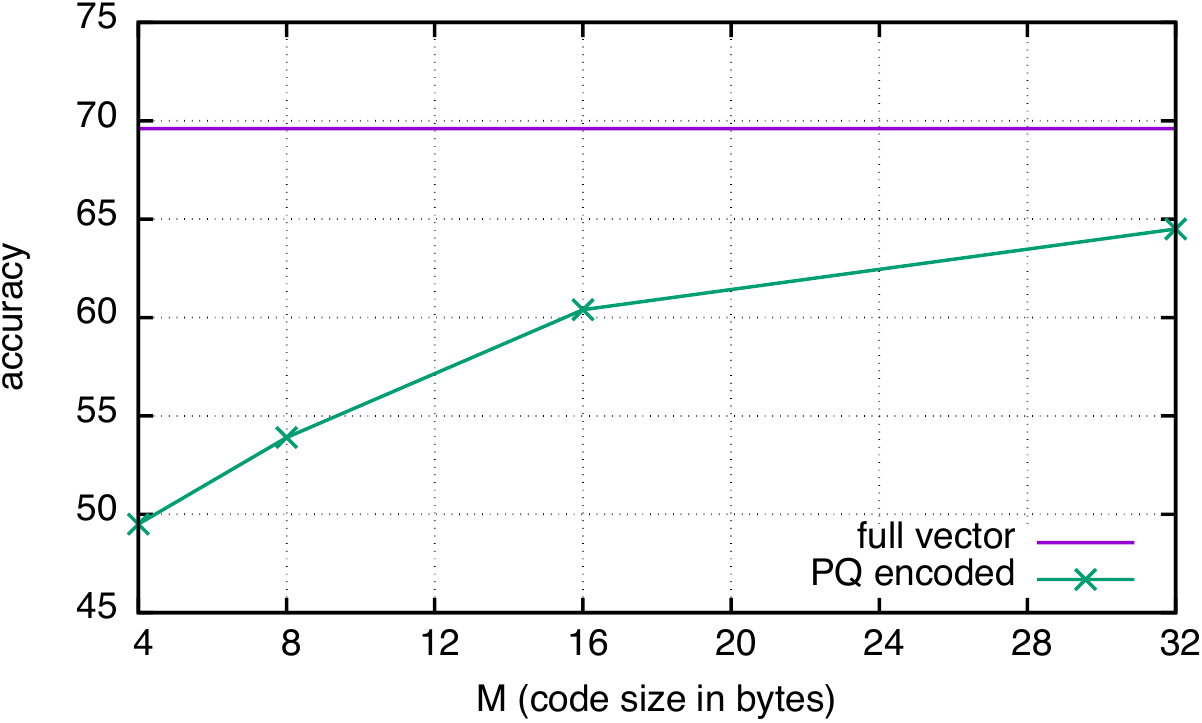}
    \end{center}
\end{figure}

We evaluate retrieval and classification methods based on hashed descriptors with our two protocols on CIFAR-100~\cite{K09}, which is the same dataset as CIFAR-10 except that it is divided into $100$ classes instead of 10), and ImageNet. 
We used the AlexNet~\cite{KSH12} architecture, with 3 (resp 5) convolutional layers (``conv") and 2 fully-connected (``fc") ones for CIFAR-100 (resp ImageNet). 

As an unsupervised baseline for hashing, we report the performance of the Product Quantizer (PQ)~\cite{JDS11}. This is an efficient method that can reconstruct approximate vectors from the codes. The PQ parameter $M$ is the number of quantizers and the number of bytes of the produced codes. 
\smallskip

\noindent\textbf{Retrieval.}
Retrieval on the classification layer is done using an inner product similarity, for all other layers we use L2 distance. 
Table \ref{tab:class_retrieval} reports the results with full activations and PQ. The performance increases monotonically with the CNN level. This is consistent with prior observations~\cite{TSF10} that attribute vectors from arbitrary classes are an efficient representation for global image matching, and further extends the findings of Section 3 to a different set of classes. The hierarchy of performance is maintained with the PQ encoding. For CIFAR-100, where the softmax outputs 75D vectors, the performance loss due to the hashing is limited even with relatively small 32-bit codes. For Imagenet, the raw retrieval performance is lower, due to the larger number of classes (250). 64-bit unsupervised encoding looses about 2\%. %
\smallskip 

\noindent\textbf{Classification.}
Results of Protocol 2 (Table \ref{tab:transfer_accuracy}) show that there is a trade-off between classification error and quantization error: activations of lower layers are more general-purpose (see, e.g., \cite{YCBL14}), so training on train25 is more effective. However, lower layers have larger activation maps, which are harder to encode, which leads to a compromise. In this example, the best transfer performance we can achieve with 4 bytes is $47.7\%$. \textbf{For a higher number of bytes, however, it is worth putting the quantization layer lower in the network}. 

Figure~\ref{fig:transfer_accuracy_conv3} shows that more bytes bring the performance closer to the full-vector performance. The margin for improvement left to supervised hashing is to bring this performance closer to the $ 69.6 \% $ obtained without any encoding.

\section{Conclusion}

In this paper, we showed that the supervised hashing protocols currently used in the literature are flawed because the evaluation is done on the same classes as the training. In this setting, encoding in binary the output of a simple classifier provides a very strong baseline. To circumvent this issue, we introduced two new protocols that evaluate hashing functions on a disjoint set of labels. The first one evaluates the retrieval performance on a disjoint set of classes. It is very close to the classical setup of unsupervised hashing, and traditional methods seem to perform well. The second protocol evaluates the accuracy of a classifier trained on hash codes with classes never seen before.

\vfill\pagebreak

\balance

\bibliographystyle{IEEEbib}
\bibliography{egbib}

\begin{thebibliography}{10}

\bibitem{NFS12}
Mohammad Norouzi, David~J Fleet, and Ruslan~R Salakhutdinov,
\newblock ``Hamming distance metric learning,''
\newblock in {\em NIPS}, 2012.

\bibitem{WLSJ13}
Jun Wang, Wei Liu, Andy~X. Sun, and Yu-Gang Jiang,
\newblock ``Learning hash codes with listwise supervision,''
\newblock in {\em ICCV}, 2013.

\bibitem{WWYL13}
Jianfeng Wang, Jingdong Wang, Nenghai Yu, and Shipeng Li,
\newblock ``Order preserving hashing for approximate nearest neighbor search,''
\newblock in {\em ACM Multimedia}, 2013.

\bibitem{ZHWT15}
Fang Zhao, Yongzhen Huang, Liang Wang, and Tieniu Tan,
\newblock ``Deep semantic ranking based hashing for multi-label image
  retrieval,''
\newblock in {\em CVPR}, June 2015.

\bibitem{KD09}
Brian Kulis and Trevor Darrell,
\newblock ``Learning to hash with binary reconstructive embeddings,''
\newblock in {\em NIPS}, December 2009.

\bibitem{WKC10}
Jun Wang, Sanjiv Kumar, and Shih-Fu Chang,
\newblock ``Semi-supervised hashing for scalable image retrieval,''
\newblock in {\em CVPR}, June 2010.

\bibitem{LWJJC12}
Wei Liu, Jun Wang, Rongrong Ji, Yu-Gang Jiang, and Shih-Fu Chang,
\newblock ``Supervised hashing with kernels.,''
\newblock in {\em CVPR}, June 2012.

\bibitem{SH09}
Ruslan Salakhutdinov and Geoffrey Hinton,
\newblock ``Semantic hashing,''
\newblock {\em IJAR}, 2009.

\bibitem{NF11}
Mohammad Norouzi and David~J. Fleet,
\newblock ``Minimal loss hashing for compact binary codes.,''
\newblock in {\em ICML}, 2011.

\bibitem{LSSHS14}
Guosheng Lin, Chunhua Shen, Qinfeng Shi, Anton van~den Hengel, and David Suter,
\newblock ``Fast supervised hashing with decision trees for high-dimensional
  data,''
\newblock in {\em CVPR}, June 2014.

\bibitem{SSLT15}
Fumin Shen, Chunhua Shen, Wei Liu, and Heng Tao~Shen,
\newblock ``Supervised discrete hashing,''
\newblock in {\em CVPR}, June 2015.

\bibitem{WZGTW16}
Xiaojuan Wang, Ting Zhang, Guo-Jun Qi, Jinhui Tang, and Jingdong Wang,
\newblock ``Supervised quantization for similarity search,''
\newblock in {\em CVPR}, June 2016.

\bibitem{DDC16}
Thanh{-}Toan Do, Anh{-}Zung Doan, and Ngai{-}Man Cheung,
\newblock ``Learning to hash with binary deep neural network,''
\newblock {\em ECCV}, 2016.

\bibitem{ZCS16}
Ziming Zhang, Yuting Chen, and Venkatesh Saligrama,
\newblock ``Efficient training of very deep neural networks for supervised
  hashing,''
\newblock in {\em CVPR}, June 2016.

\bibitem{DDNC16}
Thanh{-}Toan Do, Anh{-}Dzung Doan, Duc~Thanh Nguyen, and Ngai{-}Man Cheung,
\newblock ``Binary hashing with semidefinite relaxation and augmented
  lagrangian,''
\newblock {\em ECCV}, 2016.

\bibitem{LNH09}
C.~H. Lampert, H.~Nickisch, and S.~Harmeling,
\newblock ``Learning to detect unseen object classes by between-class attribute
  transfer,''
\newblock in {\em CVPR}, 2009.

\bibitem{TSF10}
Andrew~Fitzgibbon Lorenzo~Torresani, Martin~Szummer,
\newblock ``Efficient object category recognition using classemes,''
\newblock in {\em ECCV}, 2010.

\bibitem{R77}
Stephen~E Robertson,
\newblock ``The probability ranking principle in ir,''
\newblock {\em Journal of documentation}, 1977.

\bibitem{JFF12}
Herv\'{e} Jegou, Teddy Furon, and Jean-Jacques Fuchs,
\newblock ``Anti-sparse coding for approximate nearest neighbor search,''
\newblock in {\em ICASSP}, January 2012.

\bibitem{K09}
Alex Krizhevsky,
\newblock ``Learning multiple layers of features from tiny images,''
\newblock Tech. {R}ep., University of Toronto, 2009.

\bibitem{LWSC16}
Haomiao Liu, Ruiping Wang, Shiguang Shan, and Xilin Chen,
\newblock ``Deep supervised hashing for fast image retrieval,''
\newblock in {\em CVPR}, June 2016.

\bibitem{XPLLY14}
Rongkai Xia, Yan Pan, Hanjiang Lai, Cong Liu, and Shuicheng Yan,
\newblock ``Supervised hashing for image retrieval via image representation
  learning,''
\newblock {\em AAAI}, 2014.

\bibitem{DSLLF09}
Wei Dong, Richard Socher, Li~Li-Jia, Kai Li, and Li~Fei-Fei,
\newblock ``Imagenet: A large-scale hierarchical image database,''
\newblock in {\em CVPR}, June 2009.

\bibitem{SZ14}
K.~Simonyan and A.~Zisserman,
\newblock ``Very deep convolutional networks for large-scale image
  recognition,''
\newblock {\em arXiv preprint arXiv:1409.1556}, 2014.

\bibitem{MVPC12}
Thomas Mensink, Jakob Verbeek, Florent Perronnin, and Gabriela Csurka,
\newblock ``Metric learning for large scale image classification: Generalizing
  to new classes at near-zero cost,''
\newblock in {\em ECCV}, December 2012.

\bibitem{TE11}
A.~Torralba and A.~A. Efros,
\newblock ``Unbiased look at dataset bias,''
\newblock in {\em CVPR}, 2011.

\bibitem{BSCL14}
Artem Babenko, Anton Slesarev, Alexandr Chigorin, and Victor Lempitsky,
\newblock ``Neural codes for image retrieval,''
\newblock in {\em ECCV}, September 2014.

\bibitem{KSH12}
Alex Krizhevsky, Ilya Sutskever, and Geoffrey~E Hinton,
\newblock ``Imagenet classification with deep convolutional neural networks,''
\newblock in {\em NIPS}, December 2012.

\bibitem{JDS11}
Herv\'{e} Jegou, Matthijs Douze, and Cordelia Schmid,
\newblock ``Product quantization for nearest neighbor search,''
\newblock {\em IEEE Trans. PAMI}, January 2011.

\bibitem{YCBL14}
Jason Yosinski, Jeff Clune, Yoshua Bengio, and Hod Lipson,
\newblock ``How transferable are features in deep neural networks?,''
\newblock in {\em NIPS}, 2014.

\end{thebibliography}

\pagebreak
\nobalance

\section*{Updates since publication}
We greatly thank Svebor Karaman for detecting an error in Table 1. We corrected the error and now report performance averaged over 10 runs, along with the corresponding standard deviations. The code to reproduce the experiments on the GIST descriptors is available at \url{https://github.com/facebookresearch/supervised-hashing-baselines}.
\end{document}